\documentclass{article}
\usepackage{amsmath,graphicx,mlspconf,amsthm, multirow, makecell, url, hyperref}

\copyrightnotice{978-1-7281-6338-3/21/\$31.00 {\copyright}2021 IEEE}

\toappear{2021 IEEE International Workshop on Machine Learning for Signal Processing, Oct.\ 25--28, 2021, Gold Coast, Australia}

\theoremstyle{definition}
\newtheorem{definition}{Definition}[section]

\title{Not all noise is accounted equally: How differentially private learning benefits from large sampling rates}
\name{%
    Friedrich Dörmann$^{\dagger}$\sthanks{* The first two authors contributed equally to this work.}
    \quad Osvald Frisk$^{\dagger}$\footnotemark[1]
    \quad Lars Nørvang Andersen$^{\ddagger}$
    \quad Christian Fischer Pedersen\thanks{The authors thank the Artificial Intelligence in Rehabilitation (AIR) Project at Aarhus University for funding.}$^{\dagger}$
}
\address{%
    Aarhus University \\
    $^{\dagger}$ Department of Electrical and Computer Engineering \\%
    $^{\ddagger}$ Department of Mathematics%
}

\begin{document}

\maketitle

\begin{abstract}
Learning often involves sensitive data and as such, privacy preserving extensions to Stochastic Gradient Descent (SGD) and other machine learning algorithms have been developed using the definitions of Differential Privacy (DP). In differentially private SGD, the gradients computed at each training iteration are subject to two different types of noise. Firstly, inherent sampling noise arising from the use of minibatches. Secondly, additive Gaussian noise from the underlying mechanisms that introduce privacy. In this study, we show that these two types of noise are equivalent in their effect on the utility of private neural networks, however they are not accounted for equally in the privacy budget. Given this observation, we propose a training paradigm that shifts the proportions of noise towards less inherent and more additive noise, such that more of the overall noise can be accounted for in the privacy budget. With this paradigm, we are able to improve on the state-of-the-art in the privacy/utility tradeoff of private end-to-end CNNs.
\end{abstract}
\section{Introduction}
\label{sec:intro}
Conducting data analysis in a privacy-preserving manner has become increasingly important in recent years, as technologies for collecting sensitive data get more and more advanced. A vastly growing field in which practitioners can unintentionally leak sensitive information is deep learning. Deep learning models get ever larger in the number of parameters they employ, which can lead to overfitting and vulnerabilities to memorization of samples.
To prevent this, Differential Privacy (DP), a mathematically rigorous definition of privacy presented by Dwork in \cite{dwork2006DP}, has in recent years become the de facto standard in privacy preserving statistical algorithms, and especially in the field of machine learning.
The most prominent method of training deep models in a differentially private way is the \textit{Differentially Private Stochastic Gradient Descent} algorithm (DP-SGD) \cite{Abadi_2016}, which, as the name suggests, represents an extension to stochastic gradient descent (SGD) \cite{Bottou1991StochasticGL}.
While this algorithm has proven very effective in introducing privacy into machine learning, a downside of using it is that privately trained models still suffer a significant loss in utility (i.e. test accuracy). This is largely due the addition of Gaussian noise, which establishes privacy at the expense of gradient information in training \cite{Abadi_2016, mironov2019renyiSGM} and can prevent privately trained models from capturing the tails of underlying data distributions. However, apart from this detrimental effect that additive noise has in DP training, it has also been shown that in normal, non-private training using SGD, inherent sampling noise has clear beneficial effects. Concretely, it has been shown to act as regularization and enable the algorithm to “escape” from sharp, bad-generalizing minima and descend to “flat”, well-generalizing minima, meaning that it serves as a key component in helping models trained with SGD to generalize better \cite{keskar2017largebatch}.
\subsection{Contributions}
In this study, we investigate this dichotomy between beneficial and detrimental effects of noise with the motivation of better understanding and deconstructing the noise sources employed in private training.
We base our efforts on the idea that a great increase in the achievable privacy/utility tradeoff, i.e. model utility per privacy guarantee, could be achieved by increasing the proportion of noise that is accounted for in the privacy guarantee while simultaneously decreasing the proportion of unaccounted noise.

Given this, our contributions are as follows:
1)~We conduct an empirical comparison between additive Gaussian noise in DP-SGD and inherent sampling noise in SGD. Doing so, we show that the two types of noise can have the same beneficial effect on utility when compared to full batch gradient descent and can be used interchangeably.
2)~We provide a theoretical and practical analysis of the effect that different parameters have on the privacy guarantee when training with DP-SGD. Based on this, we show that it is possible to shift the proportions of inherent SGD noise and additive DP noise to achieve better privacy guarantees for the same overall amount of noise.
3)~With this intuition, we propose a new training paradigm for training deep learning models with differential privacy, which is focused on increasing the two privacy related hyper-parameters sampling rate ($q$) and noise multiplier ($\sigma$) substantially, to achieve the mentioned shift in noise proportions. Thus, practitioners can adopt this approach by merely changing a few parameters in training, which should immediately lead to better results in a wide range of possible applications.
4)~Using this paradigm, we present new state-of-the-art (SOTA) results in accuracy and privacy/utility tradeoffs for differentially private end-to-end CNNs trained on MNIST, Fashion-MNIST and CIFAR-10. The highlighted results shown in Table \ref{tab:Condensed_Results} display the best accuracies and privacy/utility tradeoffs achieved on each of the datasets, compared to the current SOTA.
\begin{table}[h]
    \centering
    \begin{tabular} { l|c|cc|cc| }
        \cline{2-6}
        & \multicolumn{3}{c|}{\textbf{Reference Results}} & \multicolumn{2}{c|}{\textbf{Our Results}} \\
        \cline{2-6}
        & Ref. & $\varepsilon$ & Acc.[\%] & $\varepsilon$ & Acc.[\%] \\
        \Xhline{2\arrayrulewidth}
        MNIST & \cite{papernot2020tempered} & $2.93$    & $98.1$    & ${2.93}$   &	$98.1$ $(0.11)$ \\
        \Xhline{1\arrayrulewidth}
        FMNIST & \cite{papernot2020tempered} & $2.70$   & $86.1$    & $\textbf{2.40}$   &	$\textbf{86.8}$ $\textbf{(0.17)}$ \\
        \Xhline{1\arrayrulewidth}
        CIFAR-10 & \cite{papernot2020tempered} & $7.53$ & $66.2$    & $\textbf{7.42}$   &	$\textbf{70.1}$ $\textbf{(0.20)}$ \\
        \Xhline{2\arrayrulewidth}
    \end{tabular}
    \caption{Best test accuracy achieved on the given datasets, together with the respective privacy guarantees $\varepsilon$ (lower $\varepsilon$ is better). Results are averaged over 5 runs and reported as mean and standard deviation. (See Table \ref{tab:Full_Results} for full results).
    Code available at \href{https://github.com/OsvaldFrisk/dp-not-all-noise-is-equal}{https://github.com/}\href{https://github.com/OsvaldFrisk/dp-not-all-noise-is-equal}{OsvaldFrisk/dp-not-all-noise-is-equal}.}
    \label{tab:Condensed_Results}
\end{table}
\section{Theoretical Background}
\label{sec:theory}
\subsection{Stochastic Gradient Descent}
To train deep learning models, a non-convex optimization problem of minimizing a loss function $f$ has to be solved. One of the most effective and popular optimization algorithms for tackling this problem is Stochastic Gradient Descent (SGD) \cite{Bottou1991StochasticGL}. Like Gradient Descent (GD), this method updates the parameters of a deep model iteratively over the course of training, minimizing the loss function. Each iteration updates the parameters $\theta$ of a deep model by the negative gradient of the loss function $f$ with respect to the parameters of the model, as shown in Expression \ref{eq:sgd}.
\begin{equation}
    \theta_{k+1} = \theta_k - \eta_k \left( \frac{1}{|B_k|} \sum_{i \in B_k} \nabla f_i (\theta_k) \right)
\label{eq:sgd}
\end{equation}
The difference between GD and SGD is that the former always uses the full gradient available with the given dataset, while the latter only uses approximations of the gradient, obtained by sampling a subset, a \textit{minibatch}, of the total dataset. In Expression (\ref{eq:sgd}), the $k$th minibatch is represented by $B_k$. This approximation is subject to sampling noise, meaning that SGD is approximately GD with an additional noise component added at each iteration \cite{wu2020noisy}.
\subsection{Differential Privacy in Machine Learning}
One of the most widely used definitions of Differential Privacy is ($\varepsilon, \delta$)-DP, which enables the parameterization of the privacy guarantee provided by a randomized algorithm by two variables, $\varepsilon$ and $\delta$, as can be seen in Definition \ref{ed_DP_Definition}. $\varepsilon$ expresses an upper-bound for the privacy loss of any individual sample in the dataset, known as the privacy budget. Generally, the lower $\varepsilon$, the better the privacy guarantee, as for $\varepsilon = 0$, optimal privacy would be achieved. The role of $\delta$ is to capture the probability that the bound on the privacy budget $\varepsilon$ will not hold. While allowing for such a parameter might seem unattractive it does enable much tighter composition of many privacy budgets, and when $\delta$ is kept "sub-polynomially" low, it should have no effect \cite{dworkroth2014DP}.
\begin{definition}[($\varepsilon, \delta$)-Differential Privacy \cite{dworkroth2014DP}]\label{ed_DP_Definition} A randomized algorithm $\mathcal{M}$ is ($\varepsilon, \delta$)-differentially private if for all adjacent datasets $D_1$ and $D_2$ and all $\mathcal{S} \subseteq$ Range($\mathcal{M}$)
\[Pr [\mathcal{M}(D_1) \in S] \leq \,e ^{\varepsilon} \cdot Pr [\mathcal{M}(D_2) \in S] + \delta\]
\end{definition}
In addition to ($\varepsilon, \delta$)-DP, many other definitions of DP have been proposed in recent years, most of which are relaxations of the original definition that are more tailored for specialized use like in machine learning. One of these definitions is Rényi Differential Privacy (RDP), which uses Rényi Divergence to place a bound on the log of the moment generating function of the \textit{privacy loss random variable}, providing much tighter bounds than ($\varepsilon, \delta$)-DP \cite{Mironov_2017}.

As stated, DP and especially RDP have been successfully applied in the field of deep learning \cite{Abadi_2016, bassily2014differentially}, mainly through the Differentially Private Stochastic Gradient Descent (DP-SGD) algorithm, which by now is widely adopted for private training of deep models \cite{tfprivacy, opacus}.
Concretely, to train a neural network in a differentially private manner, DP-SGD adds two steps to the training procedure. First, the gradients with respect to each sample in the dataset are clipped in their $\ell^2$-norm to bound the sensitivity of the algorithm. Second, Gaussian noise is added to the gradients, scaled to fit the previously established sensitivity, i.e. $\ell^2$-norm bound, of the clipped gradients. The variance of this noise, together with the sampling probability of the algorithm, then results in a certain ($\varepsilon, \delta$)-DP guarantee. The exact amount of this budget, that is consumed by each training iteration, is calculated and tracked by the \textit{moments accountant}, based on \cite{Abadi_2016, mironov2019renyiSGM}.
\subsection{Gradient Noise and its effects on Generalization}
\label{sec:gradientnoise}
\subsubsection{Inherent sampling noise in SGD}
Since its inception, models trained with SGD have consistently been shown to outperform models trained with GD when generalizing to a test dataset. This difference in generalization is commonly attributed to the inherent sampling noise that arises from using small minibatches of training samples rather than using the whole dataset at each step - generally speaking, the lower the batch size, the higher the scale of the sampling noise. Futhermore, these improvements in model generalization only hold up to certain batch sizes, as increasing the batch sizes beyond a certain limit again leads to a drop in test accuracy. This drop in test accuracy has been coined the "\textit{Generalization Gap Phenomenon}" \cite{keskar2017largebatch}.

One potential explanation of this phenomenon can be given when examining the inherent sampling noise more closely, and the effect that it has on SGDs convergence to local minima. Since the optimization problem is non-convex, multiple local minima are expected to be scattered throughout the loss “landscape”. Of these, it has been shown that “sharper” minima, which are characterized by a significant number of large positive eigenvalues in the hessian of the loss function $\nabla^2f(x)$, exhibit bad model generalization, while “flat” minima, which have small eigenvalues of $\nabla^2f(x)$, exhibit better generalization. It has furthermore been shown that SGD with smaller batch-sizes is more likely to converge to flat rather than sharp minima
\cite{keskar2017largebatch}.
This is due to fact that small batch SGD inherently is subject to more gradient noise than large batch SGD, which makes it more capable to escape these minima and descend to flat minima instead, leading to better test accuracy. Worded differently, noise can be a very beneficial component in yielding well generalizing models.
\begin{figure}
\begin{minipage}[b]{1.0\linewidth}
  \centering
  \centerline{\includegraphics[width=8cm]{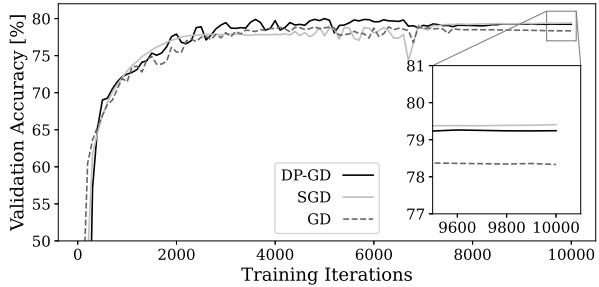}}
\end{minipage}
\caption{Generalization ability of GD, SGD and DP-GD. This experiment is a recreation of Figure 1 (a) of \cite{wu2020noisy}, extended with DP-GD. The experiment setting is the same.}
\label{fig:GDComp}
\end{figure}
\subsubsection{Additive Gaussian noise in Differential Privacy}
Despite these positive effects, too much noise can also have a detrimental effect. As suggested by Feldman in \cite{feldman2021does}, the expected loss in accuracy of differentially private models exists because of the clipping and noising of the gradients in training. In the study, the author demonstrates how private training leads to limited memorization and that memorization is crucial for achieving close-to-optimal generalization error. In other words, the perturbations introduced by differential privacy make it harder to capture the tails of the underlying distribution of the data. This leads to a degradation of gradient information at each training step, which in turn leads to a degradation in model generalization. Empirical evidence, as presented by Papernot et al. \cite{papernot2020tempered}, suggests that this effect is more significant for harder learning tasks like CIFAR-10, for which end-to-end Convolutional Neural Networks (CNN) in the current SOTA suffer a loss in test accuracy compared to non-private models of about 10\%. For easier tasks such as MNIST and Fashion-MNIST, they only show a small degradation in test accuracy of 1\% and 3\% respectively.
\section{Influences on Gradient noise}
\label{sec:noiseinfluences}
An important observation about the two types of noise presented in Section \ref{sec:gradientnoise} is that they are controlled by different parameters in training.
The inherent noise of the SGD algorithm is primarily influenced by two hyper-parameters, namely the \textit{batch size} $\text{B}$ and \textit{learning rate} $\eta$ employed in training \cite{wu2020noisy}.
On the other hand, the differentially private noise added in DP-SGD is controlled by two additional hyper-parameters, namely the \textit{clipping threshold} $C$ and the \textit{noise multiplier} $\sigma$. After the gradients are clipped to $\ell^2$-norm bound $C$ and averaged, zero-mean Gaussian noise with variance ${(C\cdot\sigma)^2}/{B^2}$ is added to the averaged gradient. The higher this variance, the better the privacy guarantee for each update.

With this established, we continue with an experimental examination of how both types of noise affect the utility of privately trained models. As mentioned in section \ref{sec:theory}, GD with the addition of noise at each iteration of training is approximately SGD. This has been shown by Wu et al. in \cite{wu2020noisy}, together with the fact that this relationship is largely independent of the class of distribution the noise is sampled from.
Furthermore, it has been suggested that the generalization gap can be closed by adding noise from another source \cite{wu2020noisy, wen2020empirical}.

Replicating and extending the experiment conducted in Figure 1 (a) of \cite{wu2020noisy} by Wu et al., we investigate the connection between the additive Gaussian noise used in training a private model using DP-SGD and the inherent sampling noise of training a non-private model using SGD, both when compared to a non-private model trained using GD, see Figure \ref{fig:GDComp}. Just as demonstrated in \cite{wu2020noisy}, SGD outperforms GD illustrating the generalization gap as expected. Consequently, we show that adding DP noise to GD has the same effect as increasing the inherent noise by lowering the batch size in SGD and enables the model to generalize just as well, demonstrating how both types of noise can be exchanged with each other.

\section{Influences on the privacy guarantee}
Given the findings in Section \ref{sec:noiseinfluences}, one can use the mentioned parameters to change the proportion of SGD-inherent noise to privacy-related noise to benefit the overall privacy budget, while retaining good generalization ability.
This does, however, require a careful examination of how both the batch size and noise multiplier affect the noise and privacy guarantee. The batch size, in the context of quantifying privacy, is expressed through the fraction $q$ called the \textit{sampling rate}. The sampling rate is calculated as the batch size divided by the training set size. Based on the so-called “privacy amplification by subsampling” principle \cite{mironov2019renyiSGM}, decreasing this ratio $q$ has a quadratic impact on lowering the privacy cost. At the same time, decreasing the noise multiplier $\sigma$ has an exponential impact on increasing the privacy guarantee as can be seen in case 1 section 3.3 in \cite{mironov2019renyiSGM} for the RDP calculation.

The privacy amplification by subsampling principle thus has had a significant impact on past and current research in differential privacy, being leveraged extensively by using low sampling rates. For example, for the results by Abadi et al. in \cite{Abadi_2016} a maximum of $q=0.04$ and for the results by Papernot et al. in \cite{papernot2020tempered}, a maximum of $q=0.034$ was used\footnote{As reported by Tramèr and Boneh in \cite{tramer2021differentially}, after recreating \cite{papernot2020tempered} based on discussions with Papernot et al.}. Furthermore, the recommended default values of the two most prominent DP libraries also does not exceed $q=0.04$ \cite{tfprivacy, opacus}. However, this prominent usage of low sampling rates also means that the scale of inherent SGD sampling noise greatly increases, which by definition is not quantified in $\varepsilon$. Furthermore, with such high sampling noise one also has to decrease $\sigma$ to not suffer in model utility. This then leads to an unappealing ratio between noise that can and cannot be accounted for by the moments accountant, being skewed to the detriment of the privacy guarantee.

Thus, we propose the following strategy. To be able to account for more of the total noise in DP-SGD training, one should utilize much higher batch sizes/sampling rates while “counter balancing” gradient noise with higher noise multipliers. With this approach, the moments account is able to quantify much more of the overall noise, leading to better privacy guarantees.
\begin{figure}
    \begin{minipage}[b]{1.0\linewidth}
      \centering
      \centerline{\includegraphics[width=8cm]{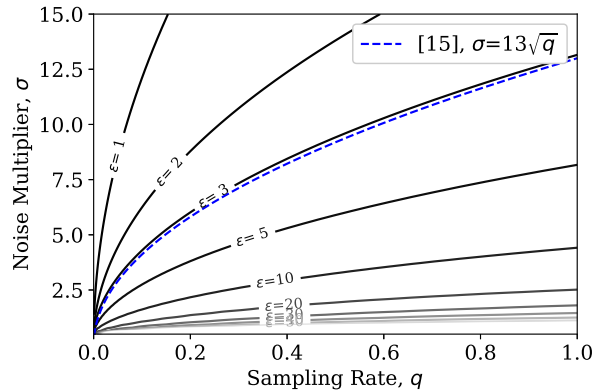}}
    \end{minipage}
    \caption{The DP noise multiplier $\sigma$ as a function of the sampling rate $q$, for different target privacy guarantees $\varepsilon$ to be reached after 60 Epochs. Additionally, the result of Tramèr and Boneh (see Fig. 6 in \cite{tramer2021differentially}) is provided for reference.}
    \label{fig:contourplot}
\end{figure}

\begin{table*}[hbt]
    \centering
    \begin{tabular} { l||l|cc||cc|cc| }
        \cline{2-8}
        & \multicolumn{3}{c||}{\textbf{Reference Results}}
        & \multicolumn{4}{c|}{\textbf{Our Results and Parameters}} \\
        \cline{2-8}
        & Reference & $\varepsilon$ & Acc. [\%] & $\varepsilon$ & Acc. [\%] & $q$ & $\sigma$ \\
        \Xhline{2\arrayrulewidth}
        \multirow{3}{*}{MNIST}
        & Feldman and Zrnic \cite{feldman2020individual}    & $1.2$     & $96.6$            & $\textbf{0.98}$   &	$\textbf{97.0}$ $\textbf{(0.19)}$ & $0.14$ & $11.85$ \\
        & Abadi et al. \cite{Abadi_2016}                    & $2.0$     & $95.0$            & $\textbf{1.95}$   &	$\textbf{97.6}$ $\textbf{(0.10)}$ & $0.17$ & $7.44$  \\
        & Papernot et al. \cite{papernot2020tempered}       & $2.93$    & $98.1$            & $2.93$            &	$98.1$ $(0.11)$               & $0.17$ & $4.65$  \\
        \Xhline{2\arrayrulewidth}
        \multirow{3}{*}{Fashion-MNIST}
        & Chen \& Lee \cite{chen2020stochastic}             & $3.0$     & $82.3$            & $\textbf{0.97}$   &	$\textbf{83.7}$ $\textbf{(0.45)}$ & $0.14$ & $11.85$ \\
        & \multirow{2}{*}{Papernot et al. \cite{papernot2020tempered}}
        & \multirow{2}{*}{$2.7$}
        & \multirow{2}{*}{$86.1$}
        & $\textbf{1.79}$               &	$86.1$ $(0.45)$  & $0.17$ & $9.23$    \\
        & & & & $\textbf{2.40}$         &	$\textbf{86.8}$ $\textbf{(0.17)}$  & $0.17$ & $6.07$    \\
        \Xhline{2\arrayrulewidth}
        \multirow{3}{*}{CIFAR-10}
        & Nasr et al. \cite{nasr2020improving}   & $3.0$     & $55.0$            & $\textbf{1.93}$   &	$\textbf{58.6}$ $\textbf{(0.38)}$ & $0.17$ & $6.60$   \\
        & \multirow{2}{*}{Papernot et al. \cite{papernot2020tempered}}
        & \multirow{2}{*}{$7.53$}
        & \multirow{2}{*}{$66.2$}
        & $\textbf{4.21}$   &	$66.2$ $(0.38)$ & $0.20$ & $4.84$    \\
        & & & & $\textbf{7.42}$   &	$\textbf{70.1}$ $\textbf{(0.20)}$ & $0.17$ & $3.12$    \\
        \Xhline{2\arrayrulewidth}
    \end{tabular}
    \caption{Test accuracies achieved for the given datasets, together with their associated privacy guarantee $\varepsilon$ (lower $\varepsilon$ is better). For each dataset, three configurations are tested and reported, together with comparable reference and SOTA results from literature. All results are averaged over 5 runs and reported as mean and standard deviation.}
    \label{tab:Full_Results}
\end{table*}

\subsection{Practial Analysis}
Given this proposal, we first verify that it is still possible to achieve good privacy guarantees using large values of $q$ and $\sigma$. To investigate this empirically, we conduct the following experiment. Over a grid of ranges $q=[10^{-3}, 1]$ and $\sigma=[0.5, 15]$, we evaluate the resulting privacy guarantee of a DP-SGD run for 60 epochs, by leveraging the privacy calculation modules provided in PyTorch Opacus \cite{opacus}. The result can be seen in Figure \ref{fig:contourplot}, displayed as a contour plot for specific values of resulting $\varepsilon$. From the plot, it is visible that the noise scale follows approximately a $\sqrt{q}$ relationship, with the slope of the function depending on the target $\varepsilon$. This is in line with findings by Tramèr and Boneh in \cite{tramer2021differentially}, in which the authors show that given a fixed privacy budget ($\varepsilon, \delta$) to be reached after $T$ epochs, the noise multiplier $\sigma$ in DP as a function of the sampling rate $q$ follows $\sigma(q) \approx c \cdot \sqrt{q}$. The authors additionally show that for the case of $(\varepsilon, \delta) = (3, 10^{-5})$ and $T=60$ Epochs, $c \approx 13$, which is displayed also in Figure \ref{fig:contourplot}.

Our analysis therefore verifies the findings in \cite{tramer2021differentially} and further expands on them by examining a range of target values for $\varepsilon$. The exponential relationship between $\sigma$ and $\varepsilon$ is apparent in the contour-lines closest to the x-axis, for which the area between contour-lines spaced 10 apart grows ever smaller. Furthermore, it is visible that the slope $c$ of the function $\sigma(q)$ for very low, i.e. good, privacy guarantees becomes very steep. This in turn means that in DP-training for a very low $\varepsilon$, $\sigma$ has to be increased more significantly, the higher the sampling rates, i.e. batch sizes.

Given these findings of the impact of different parameters on $\varepsilon$ and the knowledge of how to influence the different noise scales individually, we now proceed to give experimental results of differentially private models that are trained with the proposed paradigm of increased $q$ and $\sigma$, to investigate the utilities that such models can achieve.

\section{Experiments and Results}
\label{sec:results}
The experiments in this section are conducted on the datasets MNIST, Fashion-MNIST and CIFAR-10.
While these tasks are considered solved in much of deep learning literature, they still present a challenge to the field of differentially private deep learning, having served as the standard benchmarks for evaluating DP models in recent years, see \cite{Abadi_2016, papernot2020tempered, tramer2021differentially}.

\subsection{Experimental Setup}
All experiments are run on a workstation system using an Intel Core i9-10980XE, 64 GB of RAM and a single Nvidia RTX 3090 24 GB GPU. The implementations are mainly based on PyTorch Opacus and Weights and Biases for hyper-parameter tuning \cite{opacus, wandb}. The neural network architectures used in the experiments are adopted from Tables 1 and 2 of \cite{papernot2020tempered} to be comparable to the current SOTA.\footnote{For the exact CNN architectures used for CIFAR-10 see Table 7 in the appendix of \cite{tramer2021differentially}, which contains minor corrections from the architecture presented in the original paper \cite{papernot2020tempered}.}
The privacy budget of each of the runs is controlled by setting a target $\varepsilon$ in the privacy engine of Opacus. For each dataset, 3 experiments for 3 different, increasing levels of privacy guarantees are performed. These levels are $\varepsilon=[1, 2, 2.93]$ for MNIST, $\varepsilon=[1, 2, 2.7]$ for Fashion-MNIST and $\varepsilon=[2, 5, 7.53]$ for CIFAR-10, while $\delta=10^{-5}$ in all cases. The final levels of privacy guarantees are adopted from the current SOTA by Papernot et al. \cite{papernot2020tempered}. For each of the experiments, a hyper-parameter search is performed and the optimal configuration is finally used to generate the results. Important to point out is that in theory, hyper-parameter tuning introduces privacy leakage from the test dataset and should thus increase the privacy cost of training. However, capturing the effect in $\varepsilon$ is non-trivial and should furthermore be negligible, as pointed out in prior work \cite{tramer2021differentially}. We therefore make the same assumption (as do the mentioned reference studies) and do not account for this potential increase in $\varepsilon$.
The code to reproduce our experiments is available at
\href{https://github.com/OsvaldFrisk/dp-not-all-noise-is-equal}{https://github.com/}
\href{https://github.com/OsvaldFrisk/dp-not-all-noise-is-equal}{OsvaldFrisk/dp-not-all-noise-is-equal}.

\subsection{Results}
Table \ref{tab:Full_Results} shows the results of all the experiments conducted on the three datasets, together with the used privacy related parameters $q$ and $\sigma$, highlighting the use of the proposed high-$q$, high-$\sigma$ training paradigm. Furthermore, we compare our results against several prior SOTA results achieved with differentially private end-to-end CNNs. Like mentioned prior, these reference results are characterized to a large degree by the use of lower values for $\sigma$ and especially $q$, which in prior work generally did not exceed values of around $q=0.05$.

\begin{figure}[h]
\begin{minipage}[h]{\linewidth}
  \centering
  \centerline{\includegraphics[width=6.75cm]{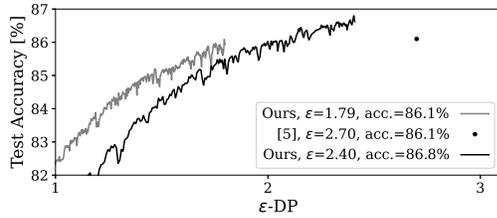}}
 \vspace{0.05cm}
  \centerline{(a) Fashion-MNIST}\medskip
\end{minipage}
\hfill
\begin{minipage}[h]{\linewidth}
  \centering
  \centerline{\includegraphics[width=6.75cm]{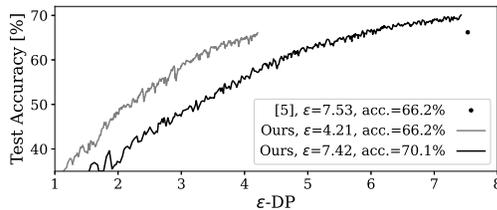}}
 \vspace{0.05cm}
  \centerline{(b) CIFAR10}\medskip
\end{minipage}
\caption{Test Accuracies as a function of the privacy budget ($\varepsilon$,$\delta=10^{-5}$)-DP, spent throughout training. Lower $\varepsilon$ is better. The results are averaged over 5 runs. The previous SOTA is provided for reference in both cases.}
\label{fig:res}
\end{figure}

As demonstrated by the results, the proposed paradigm improves the privacy/utility tradeoffs on these canonical vision benchmarks significantly. We offer better privacy guarantees while achieving as good or better test accuracies for all current SOTA results for private end-to-end CNNs. Notably, the biggest improvement is achieved on the hardest of the three tasks, CIFAR-10, for which the previous SOTA by Papernot et al. was set at 66.2\% test accuracy for $\varepsilon=7.53$, while we are able to match this accuracy at $\varepsilon=4.21$ and for $\varepsilon=7.42$ achieve a test accuracy of 70.1\%. The results are further supplemented by the plots in Figure \ref{fig:res}, which show how the test accuracy for specific runs evolves as a function of the privacy spenditure.

\section{Conclusion}
In this work, we presented a novel perspective on the relationship between the sampling rate $q$ and the noise multiplier $\sigma$ used for differentially private training of deep networks. We demonstrate that the importance of a low sampling rate, which was popularized by the “privacy amplification by subsampling” principle, is overemphasized in current literature. Our results show that one should not overestimate its effect on the privacy/utility tradeoffs when instead compared to the effect of increasing the sampling rate and noise multiplier at the same time.
As such, a novel training paradigm for training differentially private end-to-end CNNs based on these ideas of increased sampling rates was proposed, with which new state-of-the-art results in privacy/utility tradeoffs were achieved.

Our work highlights the potential of merely shifting the noise scale by using different values for fundamental parameters of DP. As such, it should be simple to transfer and apply this methodology to other private learning problems, and practitioners of differential privacy should be able to easily apply these findings to yield comparable or better privacy/utility tradeoffs in their own applications. Finally, the benefits of our work seem to especially apply on harder learning tasks for which the utility gap to non-private baselines is bigger, as we saw the largest improvement to prior work in our experiments on CIFAR-10.

\bibliographystyle{IEEEbib}
\bibliography{strings,refs}

\end{document}